\documentclass{article}
\usepackage{ismir,amsmath,cite}
\usepackage{graphicx}
\usepackage{color}
\usepackage{booktabs}
\usepackage{csquotes}
\usepackage{amsthm}
\usepackage{times}
\usepackage{soul}
\usepackage{url}
\usepackage{amsmath}
\usepackage{amsthm}
\usepackage{booktabs}
\usepackage{algorithm}
\usepackage{algorithmic}
\usepackage{amsfonts}
\usepackage{multirow}
\usepackage{multicol}
\usepackage{subfigure}
\usepackage{verbatim}
\usepackage{makecell}
\usepackage{xcolor}
\usepackage{float}

\usepackage{lineno}

\title{Content-based Controls For Music Large Language Modeling}

\multauthor
{Liwei Lin$^{1}$ \hspace{1cm} Gus Xia$^{1, 2}$
\hspace{1cm} Junyan Jiang$^{1}$
 \hspace{1cm} Yixiao Zhang$^3$} { 
  $^1$ Music X Lab, New York University Shanghai\\
$^2$ Mohamed bin Zayed University of Artificial Intelligence\\
$^3$ C4DM, Queen Mary University of London \\
}
\def\authorname{Liwei Lin, Gus Xia, Junyan Jiang and Yixiao Zhang}

\begin{document}

\maketitle
\begin{abstract}
Recent years have witnessed a rapid growth of large-scale language models in the domain of music audio. Such models enable end-to-end generation of higher-quality music, and some allow conditioned generation using text descriptions. However, the control power of text controls on music is intrinsically limited, as they can only describe music \textit{indirectly} through meta-data (such as singers and instruments) or high-level representations (such as genre and emotion). \textit{We aim to further equip the models with direct and \textbf{content-based} controls on innate music languages} such as pitch, chords and drum track. To this end, we contribute \textit{Coco-Mulla}, a \textbf{co}ntent-based \textbf{co}ntrol method for \textbf{mu}sic \textbf{l}arge \textbf{la}nguage modeling. It uses a parameter-efficient fine-tuning (PEFT) method tailored for Transformer-based audio models. Experiments show that our approach achieves high-quality music generation with \textbf{low-resource} semi-supervised learning. We fine-tune the model with less than 4$\%$ of the orignal parameters on a small dataset with fewer than 300 songs. Moreover, our approach enables effective content-based controls. We illustrate its controllability via chord and rhythm conditions, two of the most salient features of pop music. Furthermore, we show that by combining content-based controls and text descriptions, our system achieves flexible music variation generation and arrangement. Our source codes and demos are available online\footnote{\url{https://github.com/Kikyo-16/coco-mulla-repo}.}\footnote{\url{https://kikyo-16.github.io/coco-mulla/}.}.
\end{abstract}
\section{Introduction}\label{sec:introduction}

Controllable music generation encompasses the creation of music under various controls, such as musical or textual descriptions \cite{dhariwal2020jukebox, liu2023audioldm, schneider2023mo, copet2023simple}. It enables amateur users to create customized music and helps professional musicians explore new ideas for composition and arrangement. Remarkable advancements have been made in this field in recent years, particularly in text-to-music generation \cite{huang2023noise2music, agostinelli2023musiclm,schneider2023mo, copet2023simple}. These cross-modality models are trained on extensive sets of parallel text-audio data pairs, utilizing pre-trained large language models \cite{raffel2020exploring} or multi-modal embeddings \cite{guzhov2022audioclip, huang2022mulan} to establish a mapping between natural language and music. They enable high-level controls such as mood and tempo by incorporating them into a text prompt.

However, not all music information can be expressed via text. Existing models cannot yet apply effective controls on intricate musical languages (e.g., chord progressions) or directly refer to musical contents from other audio recordings. The ability to accommodate such \textit{content-based} controls is crucial for tasks such as music editing, music variation generation, and arrangement. 
For example, in AI-assistant composition systems, the generative models are required to compose based on a rough idea like motifs or counter-melodies; in music re-instrumentation, we tend to keep the underlying harmony unchanged and reinterpret the song with new instruments. 

Historically speaking, the shift from text-based (and metadata-based) models to content-based models has once happened in the realm of music information retrieval (MIR), which dramatically improved the model performances and also expanded the boundary of music understanding. In a similar fashion, we aim to push the boundary of music audio generation domain by further equipping off-shelf generative models with content-based controls. 

A simple approach to incorporating content-based control is to train a separate generative model conditioned on the provided control content \cite{copet2023simple, donahue2023singsong}. For example, both the MusicGen and MusicLM systems offer two versions of model: a vanilla text-to-music version, and a melody-conditioned version for accompaniment generation. The main issue with this simple conditioning approach lies in the high training cost, which is at the same scale as the base model in terms of both computational resources and training data. Moreover, each conditioned model can only deal with \textit{one type} of content-based control input. This rigid setting is not practical in real music production and arrangement scenarios where multiple control contents are often required for satisfactory results.

To solve the aforementioned problems, we contribute a unified approach to incorporating different content-based controls with music-audio generative models. Particularly, we see the immense potential of large-scale pre-trained models in their semantic-level understanding of music and therefore propose a novel parameter-efficient fine-tuning condition adaptor based on llama adaptor \cite{zhang2023llama}. 

The current design of the adaptor integrates the joint embeddings of symbolic music chords and piano roll and acoustic drum tracks with the pre-trained MusicGen model. In theory, the adaptor can perform on joint embeddings of any combination of content-based controls and can be integrated with any Transformer-based generative model. 
In short, the main contribution of this work is as follows:
\begin{itemize}

\item \textbf{A unified approach on content-based controls}: 
Our model enables chord and drum pattern controls via acoustic hints, achieving an arbitrary combination of textual, harmonic, and rhythmic description for the controlled generation process.

\item \textbf{Low-resource fine-tuning on pseudo-labeled datasets}: We provide a method to fine-tune a large auto-regressive audio generative model with a small-size, pseudo-labeled dataset in which all the pseudo labels are extracted using existing MIR tools. We fine-tune MusicGen \cite{copet2023simple}, an excellent text-to-music model, on 4$\%$ trainable parameters of the original model with a training set of fewer than 300 songs without text or other annotations.

\item \textbf{Flexible variation generation and arrangement}: Our model achieves flexible variation generation and arrangement of the given polyphonic piano roll by combining text prompts and content-based controls. This enables numerous downstream music-editing applications.

\end{itemize}

\section{Related work}
We review two realms of related works: 1) large-language models (LLMs) for music audio and 2) existing methods of parameter-efficient fine-tuning. 
\label{sec:format}

\subsection{Music Audio Generation}
Music audio generation necessitates extensive contextual modeling to account for the intricate structure of musical language. Recent large-scale music audio generative models, encompassing auto-regressive and diffusion-based approaches, have made remarkable strides in capturing such a long-term structure while introducing cross-modality conditions. For example, Jukebox \cite{dhariwal2020jukebox} leverages VQ-VAE \cite{van2017neural} and transformer decoders to achieve lyrics- and genre-based generation; Diffusion-based Moûsai \cite{schneider2023mo} adopts the pre-trained frozen T5 encoder \cite{raffel2020exploring} to summarize text conditions; auto-regressive MusicGen \cite{copet2023simple} realizes monophonic melody and text controls by assembling EnCodec \cite{defossez2022high}, T5 encoder, and an acoustic transformer decoder. Specifically, MusicGen is the first text- and melody-conditioned model, limited to a monophonic melody condition, and it does not accommodate drum tracks. Additionally, a contemporaneous work, Music ControNet\cite{wu2024music}, shares a similar goal with ours but is based on a diffusion model rather than on pretrained large language models (LLMs).


\begin{figure*}
 \centerline{
\includegraphics[width=\textwidth]{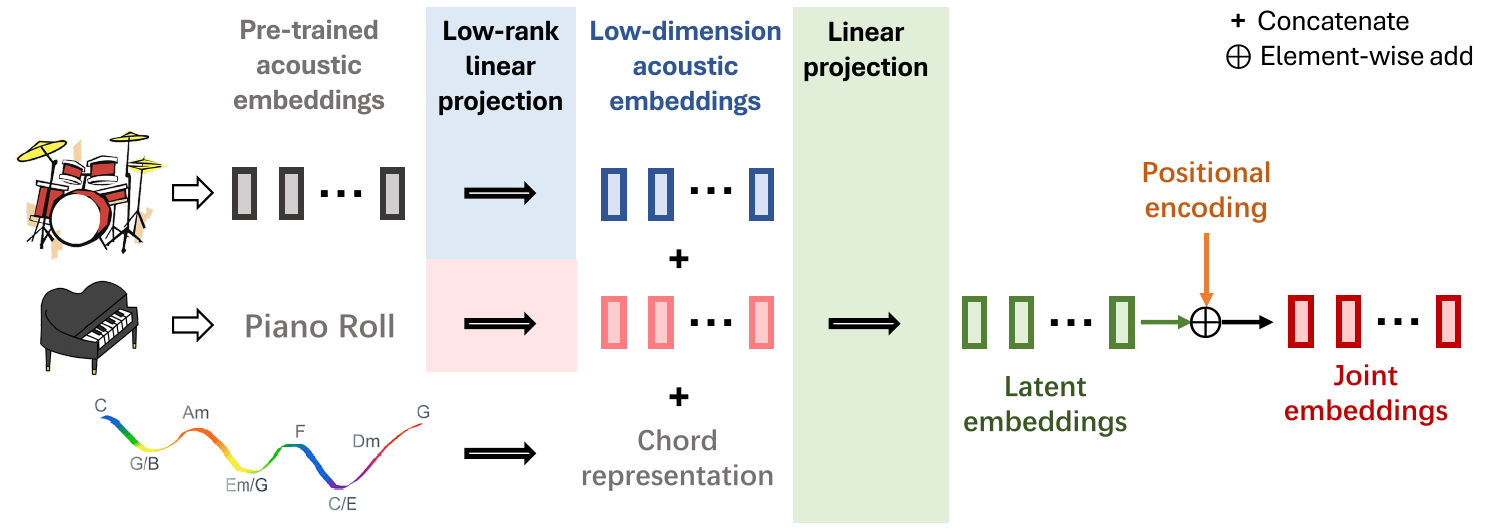}}

 \caption{The joint embedding module. We randomly mask acoustic or piano roll embedding with probability $r$ during training.}
 \label{fig:embedding}
 \vspace{.1in}
\end{figure*}

\subsection{Parameter-Efficient Fine-Tuning}
\label{ssec:peft} 
Parameter-efficient fine-tuning (PEFT) methods adapt pre-trained language models (PLMs) without fine-tuning all model parameters \cite{devlin2018bert, radford2019language}, significantly reducing computational and storage expenses. \cite{li2021prefix} and \cite{liu2021p} tailor PLMs for specific tasks by appending task-specific prefixes to input sequences. \cite{hu2021lora} employs low-rank adaption (LoRA) to fine-tune pre-trained linear transformations in PLMs. \cite{zhang2023llama} and \cite{gao2023llama} propose LLaMA-Adapter, adjusting attention outputs using prompt adaptors and zero gate scalars.  
LLaMA-Adapter also introduces a multi-modal conditional variant by incorporating \emph{global} image representations into prompt adaptors. In this study, we present a novel PEFT method, inspired by LLaMA-Adapter, designed to fine-tune large-scale models while accommodating external \emph{sequential} multi-modal conditions.

\section{Base Model}
In this work, we choose MusicGen \cite{copet2023simple}, an excellent Transformer-based music audio language model, as our base model. MusicGen offers two variants: a text-only model and a melody-based model. The melody-based model conditions its generation on the dominant time-frequency bin of the audio chromagram and text prompts, limiting it to monophonic conditioning and preventing it from incorporating rhythmic drum patterns. In this study, we adopt the text-only MusicGen as our base model, augmenting it with content-based controls via the proposed PEFT method.  

The largest text-only MusicGen consists of 3 components: a pre-trained EnCodec, a pre-trained T5 encoder, and an acoustic transformer decoder. The transformer decoder comprises $N=48$ layers, each including a causal self-attention block and a cross-attention block to handle condition text prompts. 

MusicGen tokenizes audio signals using EnCodec \cite{defossez2022high}, a Residual Vector Quantization (RVQ) \cite{zeghidour2021soundstream} auto-encoder, compressing signals of sample rate $S_r = 32000$ into discrete codes of a low frame rate $f_s=50$. The acoustic transformer decoder takes these tokens as its input.

\section{Methodology}
\label{gen_inst}

Our approach consists of 2 components: 1) a joint embedding encoder to integrate content-based controls, and 2) a condition adaptor to fine-tune MusicGen by incorporating the learned joint embeddings. To maintain the ability of the vanilla MusicGen to associate text with music, we train the adapter for only the self-attention blocks of the acoustic transformer decoder. During training, all the parameters of MusicGen are frozen.

\subsection{Joint Symbolic and Acoustic Embedding}
\label{ssec:Joint}
To incorporate the desired chord progression and to borrow musical content from another audio recording simultaneously, we design a joint symbolic and acoustic embedding. For precise alignment of generated music with acoustic hints, we use a frame-wise representation for both symbolic and acoustic data.

\begin{table}[t]\small
\begin{center}
  \begin{tabular}{  c | c |c |c | c }
    \toprule
    \textbf{Chord Name} &\textbf{Pitch} & \textbf{Root} & \textbf{Bass} & \textbf{Pitch Indices} \\ \midrule
    C:maj&C, E, G & C (0) & C (0)&0, 4, 7 \\ \midrule
    C:maj/E&C, E, G & C (0) & E (4)&0, 4, 7 \\ \midrule
    D:min&D, F, A & D (2) & D (2)&0, 3, 7 \\
    \bottomrule
  \end{tabular}
\caption{\label{table:chord}Symbolic chord data structure.}
\end{center}
\end{table}

\subsubsection{Symbolic Chord and MIDI Representation}
We describe a chord as a combination of a root pitch class, the bass pitch class, and the chroma representation of the chord quality. As shown in Table~{\ref{table:chord}}, we represent a chord as $\{\mathrm{root}, \mathrm{bass}, \bm{m}\}$ ($\mathrm{root}, \mathrm{bass}\in\{0, 1, ..., 11\}$), with $\bm{m}\in\mathbb{R}^{12}$ being a multi-hot positional vector indicating the active pitches in the octave starting from the root note. Define $\bm{c}_i\in\mathbb{R}^{12+12+12+1}$ to represent the $i^{\rm {th}}$-frame chord: 
\begin{equation}
\bm{c}_i=
    \begin{cases}
    [ e(\mathrm{root}); e(\mathrm{bass}); \bm{m}; 0]
  ,& \text{ if }i^{\rm {th}}\text{ frame has a chord } \\ 
 [ \bm{0}; \bm{0}; \bm{0}; 1],& \text{ otherwise}
\end{cases},
\end{equation}
where $e$ is a function from an index $j\in \{0, 1, ..., 11\}$ to its one-hot vector $e(j)\in \mathbb{R}^{12}$.

We represent MIDI using the piano roll format. Assume $\bm{p}_i \in \{0, 1\}^{128}$ is the $i^{\rm th}$ frame in the piano roll indicating the presence of each pitch. We further compress the sparse piano roll into a low-dimension MIDI representation $\bm{p}'_{i}$ using a trainable matrix $\bm{W}_{\rm p}\in\mathbb{R}^{d_1\times 128}$:
\begin{equation}
    \bm{p}'_{i} =\bm{W}_{\rm p}^{\mathsf{T}}\bm{p}_{i}\in\mathbb{R}^{d_1}.
\end{equation}

Throughout the training process, we use pseudo chord annotations obtained through a chord recognition model from \cite{jiang2019large} and MIDI annotations via an automatic music transcription model from \cite{wu2021omnizart}.

\subsubsection{Acoustic Representation}
We convert the separated drum stem to discrete codes using EnCodec \cite{defossez2022high}.
Instead of directly modeling these discrete codes, we pass the $i^{\textrm{th}}$-frame codes through the frozen input embedding layer of MusicGen transformer decoder to obtain a pre-trained acoustic embedding $\bm{h}_{i}\in\mathbb{R}^{2048}$. Such a continuous pre-trained representation is more robust since it can address the issue of utilizing discrete codes not present in the training data during the inference stage.

To mitigate overfitting and reduce training complexity, we employ a trainable low-rank matrix $\bm{W}_\mathrm{a} \in\mathbb{R}^{2048\times d_2}$ to map $\bm{h}_{i}$ into a lower-dimensional space:
\begin{equation}
    \bm{h}'_{i} = \bm{W}_\mathrm{a} ^\mathsf{T}\bm{h}_{i}\in\mathbb{R}^{d_2}.
\end{equation}

We set $d_2=d_1=12$ in our experiments. 

\begin{figure*}[htbp]
    \centering
    \begin{minipage}[b]{0.5\textwidth}
        \centering
        \subfigure[To incorporate content-based control from the joint embeddings, the condition prefix joins into the last $L$ frozen self-attention layers of the MusicGen decoder.]{
            \includegraphics[width=\linewidth]{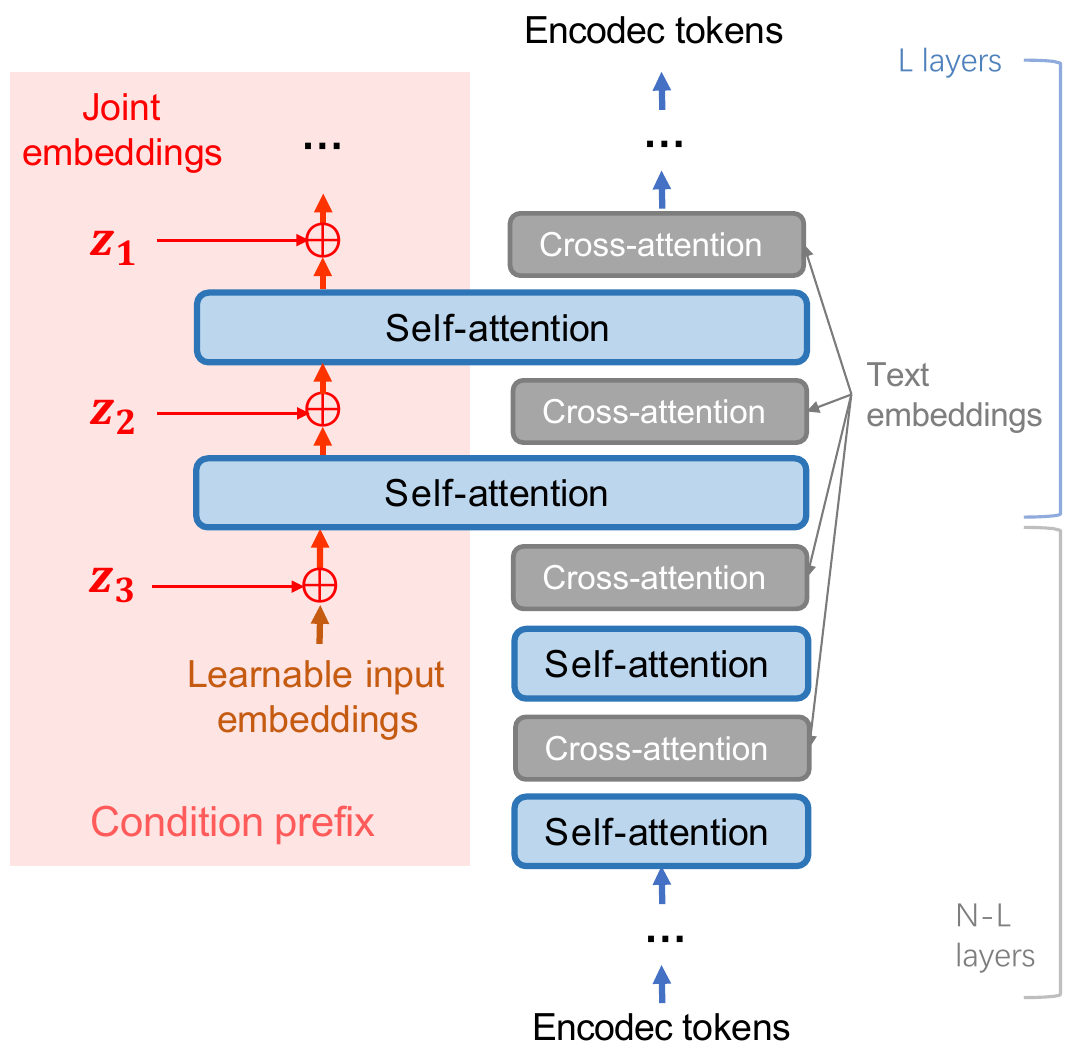}
            \label{fig:big_image}
        }
    \end{minipage}%
    \hskip 0.3in
    \begin{minipage}[b]{0.4\textwidth}
        \centering
        \subfigure[Inside the condition prefix, each position can attend to all other positions within the condition prefix.]{
            \includegraphics[width=\linewidth]{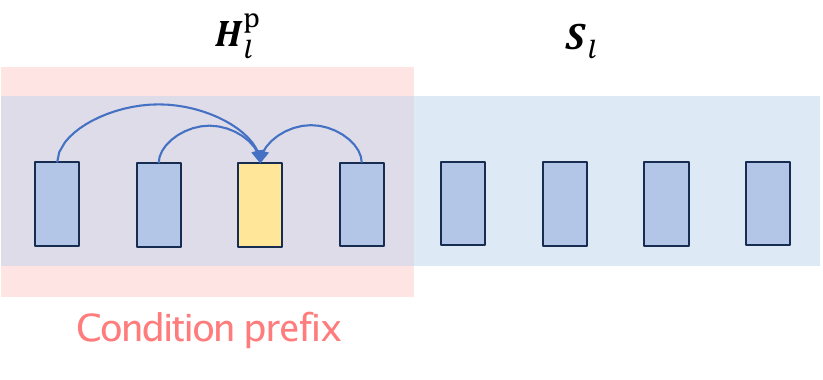}
            \label{fig:adaptors-r}
        }
        \vfill
        \vskip 0.2in
        \subfigure[Outside the condition prefix are the hidden embeddings corresponding to Encodec tokens. They attend to all preceding positions (following the causal mask) and the entire condition prefix.]{
            \includegraphics[width=\linewidth]{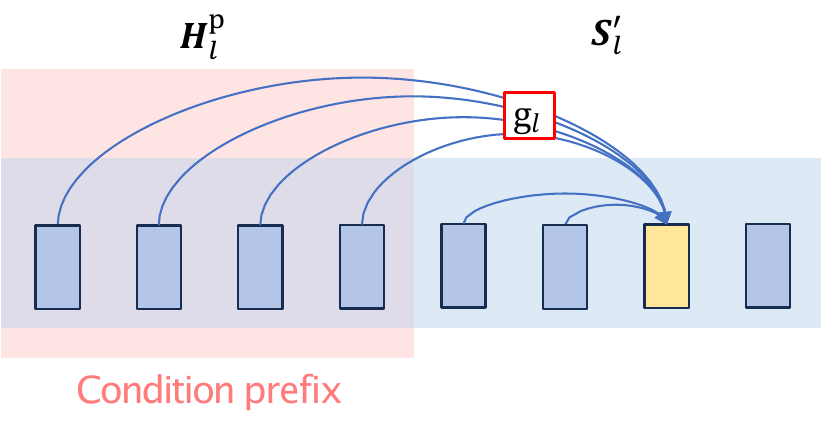}
            \label{fig:adaptors-x}
        }
    \end{minipage}
    \caption{Condition adaptor. The condition prefix is injected to the self-attention mechanism of the MusicGen transformer decoder. All transformation matrices in MusicGen are frozen. Only the input embeddings, joint embedding encoders, and the gate factors are trainable.}
    \vspace{.1in}
  \label{fig:adaptors}
\end{figure*}

\subsubsection{Masking Scheme and Positional Encoding}
During training, we randomly mask MIDI and acoustic representation with a probability $r$ independently: 
\begin{equation}
   \bm{z}^{\rm p}_{i}=
    \begin{cases}
    \bm{p}'_{i},& \text{ if not masked} \\ 
 \bm{s}^{\rm p}_{i},& \text{ otherwise}
\end{cases},
\end{equation}
\begin{equation}
   \bm{z}^{\rm a}_{i}=
    \begin{cases}
    \bm{h}'_{i},& \text{ if not masked} \\ 
 \bm{s}^{\rm a}_{i},& \text{ otherwise}
\end{cases}.
\end{equation}

Here, $\bm{s}^{\rm p}_i$ and $\bm{s}^{\rm a}_{i}$ are learnable masked embeddings. The masking strategy trains the model to follow an arbitrary combination of conditional tracks during inference. We set $r=0.4$ in our experiments.

We introduce a learnable matrix $\bm{W}_{\rm e}\in\mathbb{R}^{(d_1+d_2+37)\times d}$ to incorporate the above embeddings and a learnable positional embedding $\bm{z}^{\rm pos}_i\in\mathbb{R}^{d_1 + d_2 + 37}$ to facilitate sequential modeling. The final joint symbolic and acoustic embedding is as follows:
\begin{equation}
    \bm{z}_i = \bm{W}_{\rm e}^{\mathsf{T}}([\bm{c}_i; \bm{z}^{\rm p}_i; \bm{z}^{\rm a}_i] + \bm{z}^{\rm pos}_i) \in \mathbb{R}^{d}.
\end{equation}

Finally, let $T$ be the total number of frames. The complete sequential joint embedding is then:
\begin{equation}
\label{eq:z}
    \bm{z} =\{\bm{z}_1, \bm{z}_2, ..., \bm{z}_T\}\in\mathbb{R}^{T\times d}.
\end{equation}

\subsection{Condition Adaptor}
\label{ssec:condition-adaptor}
To plug the joint embeddings into MuiscGen, we present a novel condition adaptor that can take time-varying sequential conditions. In the vanilla Transformer, each self-attention layer operates on a sequence of $T$ hidden embeddings corresponding to $T$ frames of Encodec tokens. In the proposed condition adaptor, as shown in Fig~\ref{fig:adaptors}, for the last $L$ layers of the MusicGen decoder, we expand the sequence of hidden embeddings to $2T$, where $T$ new positions take on the task of incorporating and processing condition-related information. We call the newly introduced positions the \textit{condition prefix}.

Specifically, we insert a sequence of learnable input embeddings into the $(N-L+1)^\textrm{th}$ MusicGen transformer decoder layer, initiating the condition prefix. Inside the condition prefix, we pass the hidden states only through self-attention layers, skipping cross-attention layers. 

Let $\bm{H}^{\rm p}_{l} \in\mathbb{R}^{T\times d}$ ($N-L+1 \leq l \leq N$) represent the output of the $l^\textrm{th}$-layer attention layer for the condition prefix. $\bm{H}^{\rm p}_{0}$ is a sequence of learnable input embeddings. We compute the condition prefix as follows:
\begin{equation}
    \bm{Q}^{\rm p}_l, \bm{K}^{\rm p}_l, \bm{V}^{\rm p}_l
    = \text{QKV-projector}({\bm H}^{\rm p}_l + \bm{Z}_{l}),
\end{equation}
\begin{equation}
    {\bm H}^{\rm p}_{l + 1}
    = \text{Self-Attention}(\bm{Q}^{\rm p}_l, \bm{K}^{\rm p}_l, \bm{V}^{\rm p}_l),
\end{equation}
where the sequential joint embeddings $\bm{Z}_l$ is defined in Eq~(\ref{eq:z}). Note that we learn distinct joint embeddings for each decoder layer. Since the adaptor aims at capturing the long-term contextual information of the sequential joint embeddings, it \emph{does not} employ a causal attention mask for the condition prefix. Additionally, the condition prefix does not attend to the Encodec token frames, as shown in Fig~\ref{fig:adaptors-r}.

\begin{table*}[ht]
 \begin{center}
 \setlength{
\tabcolsep}{2.mm}{
 \begin{tabular}{c|ccc|ccc}
\Xhline{3\arrayrulewidth}
&\textbf{Chord$_{\rm rec}$ $\uparrow$}&\textbf{Chord$^{*}_{\rm rec}$ $\uparrow$}&\textbf{Beat$_{F_1} \uparrow$}&\textbf{CLAP$_{\rm scr}\uparrow$}&\textbf{FAD}$^{*}_{\rm vgg}\downarrow$&\textbf{FAD}$_{\rm vgg}\downarrow$\\
\hline
\textbf{Chord-only} &0.412 & 0.195 & - & \textbf{0.401}&6.209&6.695 \\
\textbf{MIDI-only} &0.649 & 0.406 & - & 0.381 & 7.105&7.094 \\
\textbf{Drums-only} &0.530 & 0.267 & 0.856 & 0.360 & 3.845& 4.933\\
\textbf{Full} &\textbf{0.791} & \textbf{0.524} & \textbf{0.864}& 0.351 & \textbf{3.697}& \textbf{4.370}\\
\hline 
\textbf{MusicGen} & - & - & - & 0.441&6.434&6.847
\\
\textbf{Oracle} &0.885 & 0.695 & 0.898 & - & -
\\
\Xhline{3\arrayrulewidth}
 \end{tabular}}
\end{center}
 \caption{The performance of the model with $L=48$ on RWC-POP-100 subset. Oracle scores gauge the performance of the chord recognition model \cite{jiang2019large} and beat tracking model \cite{bock2016madmom} on the ground-truth audio.}
 \label{table:exp1}
 \vspace{.1in}
\end{table*}

\begin{table*}[t]
 \begin{center}
 \setlength{
\tabcolsep}{2.6mm}{
 \begin{tabular}{c|cc|cc|cc}
\Xhline{3\arrayrulewidth}
&\multirow{2}{*}{\textbf{Total}}&\multirow{2}{*}{\textbf{Trainable}}&\multicolumn{2}{c|}{\textbf{CLAP}$^{*}_{\rm scr}\uparrow$}&\multicolumn{2}{c}{\textbf{Chord}$_{\rm rec}\uparrow$}\\
\cline{4-7}
$L$&&&\textbf{Chord-only}&\textbf{Full}&\textbf{Chord-only}&\textbf{Full}
\\
  \hline
\textbf{12}
&3.29B&$0.87\%$&\textbf{0.428}&\textbf{0.371}&0.239&0.672
\\
\textbf{24}
&3.31B&$1.66\%$&0.408&0.358&0.397&0.747
\\
\textbf{36}
&3.33B&$2.44\%$&0.396&0.344&0.410&0.772
\\
\textbf{48}
&3.36B&$3.20\%$&0.401&0.351&\textbf{0.412}&\textbf{0.791}
\\
\Xhline{3\arrayrulewidth}
 \end{tabular}}
\end{center}
 \caption{The performance of models with different $L$ values under the chord-only condition.}
 \label{table:exp3}
  \vspace{.1in}
\end{table*}

For the non-prefix part, hidden states are passed through both self-attention and cross-attention layers. Let $\bm{H}_{l}\in\mathbb{R}^{T\times d}$ ($1 \leq l \leq N$) represent the output of the $l^\textrm{th}$ attention layer for the encoded tokens. we compute vanilla attention output ${\bm S}_{l}$ as follows:
\begin{equation}
    \bm{Q}_l, \bm{K}_l, \bm{V}_l
    = \text{QKV-projector}({\bm H}_l),\label{eq:self-attn}
\end{equation}
\begin{equation}
    {\bm S}_{l}
    = \text{Self-Attention}(\bm{Q}_l, \bm{K}_l, \bm{V}_l).\label{eq:self-attn}
\end{equation}
To incorporate condition information, in the last $L$ layers, we compute cross attention $\bm{S}^{'}_{l}$ between $\bm{Q}_l$ and $\{\bm{K}^{\rm p}_l, \bm{V}^{\rm p}_l\}$. We leverage Self-Attetion layers to compute $\bm{S}^{'}_{l}$ rather than Cross-Attention layers since the controls are closer to the audio modality than the textual modality. To make \{query, key, value\} more compatible, we use the fusion of $\bm{Q}_l$ and $\bm{Q}^{\rm p}_l$ as the query instead of a single $\bm{Q}_l$:
\begin{equation}
    {\bm S}^{'}_{l}
    = \text{Self-Attention}(\bm{Q}_l + \bm{Q}^{\rm p}_l, \bm{K}^{\rm p}_l, \bm{V}^{\rm p}_l).\label{eq:self-attn}
\end{equation}
Following this, we combine ${\bm S}_{l}$ and ${\bm S}^{'}_{l}$ using a zero-initialized learnable gating factor $g_l$. Additionally, to maintain the text controllability of the model, we then compute the cross attention between textual embedding and them:
\begin{equation}
    \bm{H}_{l + 1}  =  \text {Cross-Attention}({\bm S}_{l} + g_l \cdot {\bm S}^{'}_{l}, text).
\end{equation}


\begin{figure*}[ht]
  \centering
{\includegraphics[width=\columnwidth]{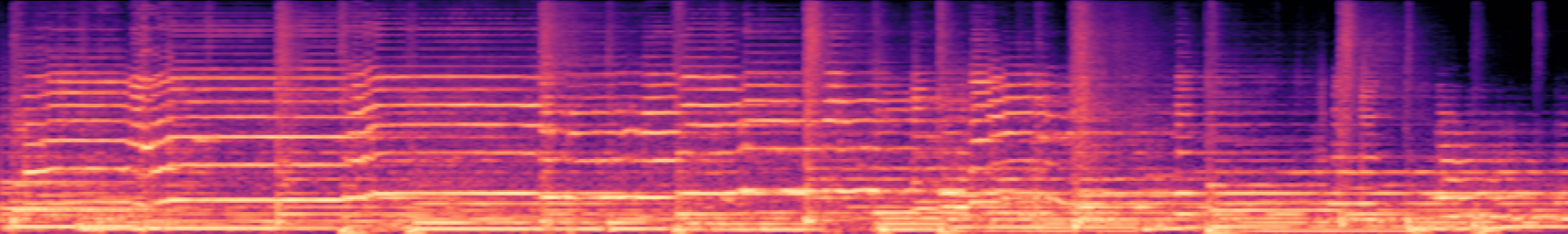}}
\hskip 0.15in
{\includegraphics[width=\columnwidth]{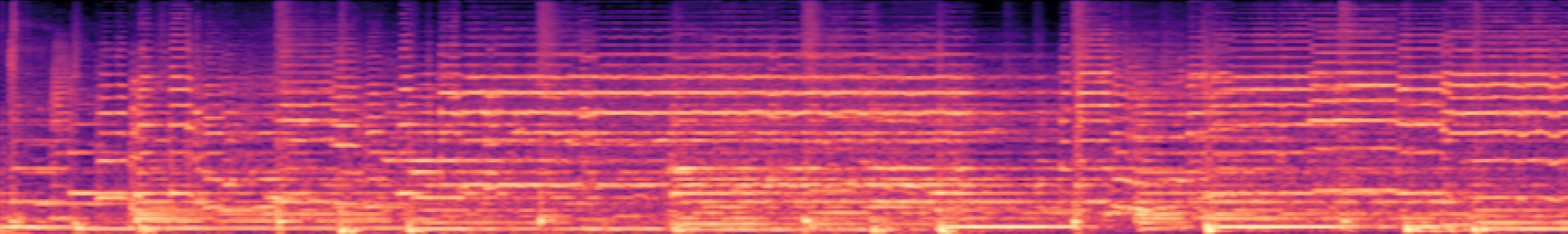}}
\vskip 0.1in
{\includegraphics[width=\columnwidth]{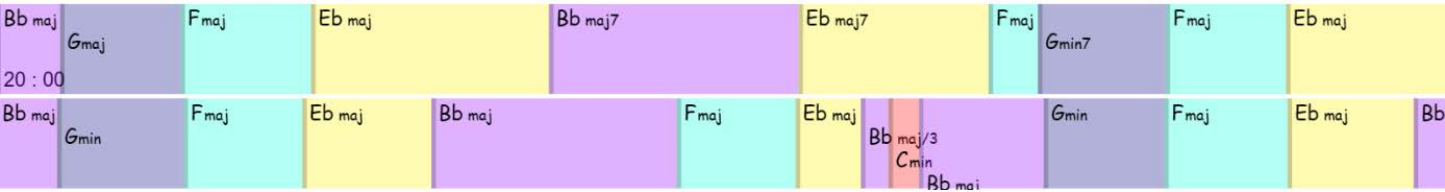}}
\hskip 0.15in
{\includegraphics[width=\columnwidth]{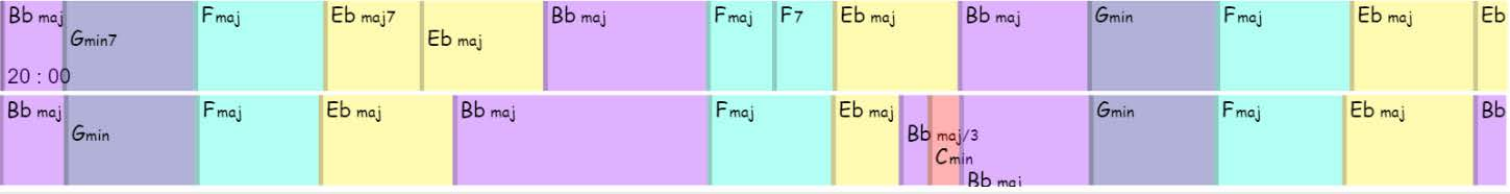}}
\subfigure[Chord and rhythm control]
{\label{fig:spec1:3a}\includegraphics[width=\columnwidth]{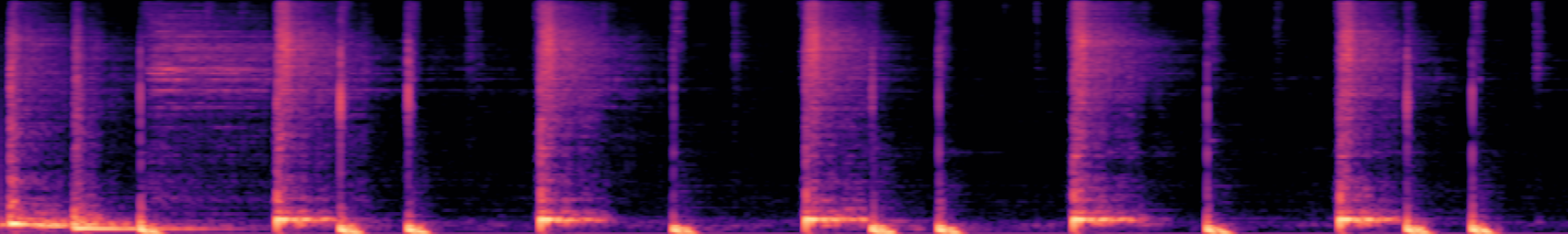}}
\hskip 0.15in
\subfigure[Chord and rhythm control with MIDI hints]
{\label{fig:spec1:3b}\includegraphics[width=\columnwidth]{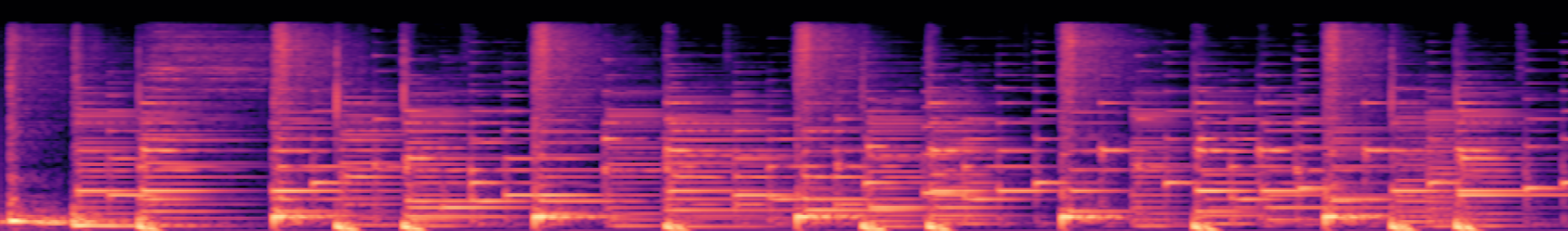}}
  \caption{Comparison of generated samples and groundtruth. The top two rows are generated samples, while the bottom rows are reference soundtracks. The text prompt is \emph{``lazy jazz composition features a captivating saxophone solo that effortlessly melds with piano chords, skillfully weaving its way through the melody with languid grace. Instruments: saxophone, piano, drums"}.}
  \label{fig:spec1}
\end{figure*}
\begin{figure}
    \centering
    \includegraphics[width=\columnwidth]{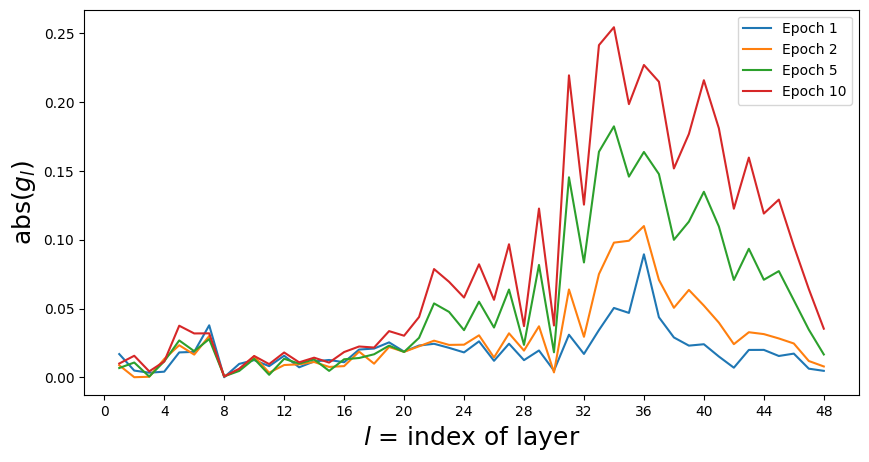}
    \caption{The variation of $|g_l|$ during training.}
            \label{fig:spec4}
\end{figure}

All the layers in MusicGen are frozen, including the QKV-projector, Self-Attention, and Cross-Attention layers. Hence, the total trainable parameters only comprise $\bm{H}^{\rm p}_0$, $\bm{W}_{\rm p}$, $\bm{W}_{\rm a}$, $\bm{W}_{\rm e}$, $\bm{z}^{\rm pos}$, and $g_l$. Moreover, the proposed adaptor can learn in a semi-supervised manner using pseudo-separated tracks, pseudo MIDI, and pseudo chord labels. Additionally, during training, each music piece is assigned to a vague text description randomly sampled from a small set of predefined phrases, eliminating the requirement for text-audio data pairs.

\section{Experiment}
\label{sec:exp}
\subsection{Datasets}

 The training dataset consists of 299 \emph{unannotated instrumental} songs. We collect 150 of them from an open-source dataset MUSDB18 \cite{rafii2017musdb18} and download the remain 149 songs from the internet. The latter subset predominantly consists of Pop songs, with a limited data of other genres such as Jazz and Rock. We omit the silent start and end segments of each training song, yielding 17.12 hours of audio. We employ Demucs to extract drum tracks, a chord recognition model \cite{jiang2019large}, an automatic transcription model \cite{Wu2021}, and a beat tracking model \cite{bock2016madmom} to generate pseudo chord, MIDI, and beat labels.

 The test set comprises 50 songs with chord, beat, and MIDI annotations from RWC-POP-100 \cite{goto2002rwc}. Vocals are excluded by a music source separation model Demucs \cite{defossez2019demucs}. 


\subsection{Training Configuration}
We train the proposed model using 4 RTX8000s with an initial learning rate of 2e-3 and a batch of 24 20-second samples for 10 epochs. We set the warm-up epoch to 2 and update the model using a cross-entropy reconstruction loss. During training, for each audio sample, we simply sample a text prompt from a text description set for each music segment: \emph{\{melodic music, catchy song, a song, music tracks\}}. 

\subsection{Evaluation}
We separate each audio clip in the test set into 4 stems using Demucs and discard the vocal track. As shown in Table~\ref{table:exp1}, \emph{``Chord-only"} signifies no drums and MIDI controls, while \emph{``Full"} indicates both drums and MIDI controls. Within each group, we generate 16 20-second audio samples per given chord progression, employing various text prompts while keeping the same chord unchanged. In total, we have 4 test groups, each with $800$ generated audio samples.

We report weighted recall score for chord accuracy, standard F-measure for rhythm control, CLAP \cite{wu2023large} score for text control evaluation, and Fréchet Audio Distance (FAD) \cite{kilgour2018fr} for audio quality measure. As depicted in Table~\ref{table:exp1}, Chord$_{\rm rec}$ represents chord root accuracy, and Chord$^{*}_{\rm rec}$ assesses full chord accuracy. FAD$^{*}_{\rm vgg}$ quantifies the audio dissimilarity between generated samples and groundtruth audio from the RWC-POP-100 subset, whereas FAD$_{\rm vgg}$ assesses this dissimilarity using the remaining 50 audios within RWC-POP-100.



\subsection{Results}
As illustrated in Table~\ref{table:exp1} and Figure~\ref{fig:spec1}, our model excels in chord and rhythm control while maintaining the text-conditioned ability, even though we do not train the model with real text annotations. We do not report chord and beat accuracy for the baseline model, as it cannot use these controls. Code details and more demos are publicly available online.

\subsubsection{Low-resource Fine-tuning}

During fine-tuning, our observations suggest that our model works better with a smaller training dataset characterized by high-quality audio fidelity than a larger one featuring pseudo-separated instrumental ground truth. As indicated in Table~\ref{table:exp3}, we discern a trade-off between controllability and semantic correlation. As the number of trainable layers increases, the model achieves a simultaneous improvement in chord recall score while witnessing a reduction in CLAP$_{\rm src}$. In addition, as shown in Fig~\ref{fig:spec4}, as the layers go deeper, the absolute value of gate factor $g_l$ increases. It demonstrates the proposed adapter primarily affects the topmost layers of the decoder transformer, implying that the lower layers are likely responsible for modeling high-level semantics, while the upper layers shape the finer details of the content. 

\subsubsection{Chord and Rhythm Control}

As shown in Table~\ref{table:exp1} and Fig~\ref{fig:spec1:3a},
 the chord control capability of our model strengthens as MIDI hints are provided. Additionally, the results highlight the model's adeptness at rhythm control when a drum track is included. However, in cases of semantic conflict between the content-based condition and the text prompt, we observe that the model tends to prioritize the former, leading to music that aligns with the drum pattern rather than the prompt.

\subsubsection{Variation Generation and Arrangement}

As depicted in Fig~\ref{fig:spec1:3b}, our model, with the assistance of MIDI hints, can produce variations by integrating musical elements from conditioned MIDI tracks, such as motifs and walking bass. Furthermore, we have observed instances where the generated audio aligns with the original main or counter melodies. This facilitates idea-driven variation generation and semantic-based arrangement within the provided polyphonic piano roll.

\section{Conclusion}
\label{sec:minhead}
We present a content-based music generative model, achieved by fine-tuning a pre-trained Transformer-based audio language model using the proposed condition adaptor. Our experimental results substantiate its proficiency in seamlessly integrating chord progressions, rhythm patterns, MIDI, and text prompts into the generated music. Our work bridges the gap of direct control via musical elements and audio conditions in the music audio generation field. Furthermore, the proposed condition adaptor facilitates efficient low-resource fine-tuning, even with a relatively small unannotated training set.
Nonetheless, results generated with conflicting audio, MIDI, and text prompts may lack musicality and may not fully meet semantic control expectations. In the future, we aim to explore further enhancements in the areas of harmonic direct control and content-based generation.

\section{Acknowledgments}
This work was supported in part through the NYU and NYUSH IT High Performance Computing resources, services, and staff expertise.
\bibliography{ISMIRtemplate}

\begin{thebibliography}{10}
\providecommand{\url}[1]{#1}
\csname url@samestyle\endcsname
\providecommand{\newblock}{\relax}
\providecommand{\bibinfo}[2]{#2}
\providecommand{\BIBentrySTDinterwordspacing}{\spaceskip=0pt\relax}
\providecommand{\BIBentryALTinterwordstretchfactor}{4}
\providecommand{\BIBentryALTinterwordspacing}{\spaceskip=\fontdimen2\font plus
\BIBentryALTinterwordstretchfactor\fontdimen3\font minus \fontdimen4\font\relax}
\providecommand{\BIBforeignlanguage}[2]{{%
\expandafter\ifx\csname l@#1\endcsname\relax
\typeout{** WARNING: IEEEtran.bst: No hyphenation pattern has been}%
\typeout{** loaded for the language `#1'. Using the pattern for}%
\typeout{** the default language instead.}%
\else
\language=\csname l@#1\endcsname
\fi
#2}}
\providecommand{\BIBdecl}{\relax}
\BIBdecl

\bibitem{dhariwal2020jukebox}
P.~Dhariwal, H.~Jun, C.~Payne, J.~W. Kim, A.~Radford, and I.~Sutskever, ``Jukebox: A generative model for music,'' \emph{arXiv preprint arXiv:2005.00341}, 2020.

\bibitem{liu2023audioldm}
H.~Liu, Z.~Chen, Y.~Yuan, X.~Mei, X.~Liu, D.~Mandic, W.~Wang, and M.~D. Plumbley, ``Audioldm: Text-to-audio generation with latent diffusion models,'' \emph{arXiv preprint arXiv:2301.12503}, 2023.

\bibitem{schneider2023mo}
F.~Schneider, Z.~Jin, and B.~Sch{\"o}lkopf, ``Mo\^{u}sai: Text-to-music generation with long-context latent diffusion,'' \emph{arXiv preprint arXiv:2301.11757}, 2023.

\bibitem{copet2023simple}
J.~Copet, F.~Kreuk, I.~Gat, T.~Remez, D.~Kant, G.~Synnaeve, Y.~Adi, and A.~D{\'e}fossez, ``Simple and controllable music generation,'' \emph{arXiv preprint arXiv:2306.05284}, 2023.

\bibitem{huang2023noise2music}
Q.~Huang, D.~S. Park, T.~Wang, T.~I. Denk, A.~Ly, N.~Chen, Z.~Zhang, Z.~Zhang, J.~Yu, C.~Frank \emph{et~al.}, ``Noise2music: Text-conditioned music generation with diffusion models,'' \emph{arXiv preprint arXiv:2302.03917}, 2023.

\bibitem{agostinelli2023musiclm}
A.~Agostinelli, T.~I. Denk, Z.~Borsos, J.~Engel, M.~Verzetti, A.~Caillon, Q.~Huang, A.~Jansen, A.~Roberts, M.~Tagliasacchi \emph{et~al.}, ``Music{LM}: Generating music from text,'' \emph{arXiv preprint arXiv:2301.11325}, 2023.

\bibitem{raffel2020exploring}
C.~Raffel, N.~Shazeer, A.~Roberts, K.~Lee, S.~Narang, M.~Matena, Y.~Zhou, W.~Li, and P.~J. Liu, ``Exploring the limits of transfer learning with a unified text-to-text transformer,'' \emph{The Journal of Machine Learning Research}, vol.~21, no.~1, pp. 5485--5551, 2020.

\bibitem{guzhov2022audioclip}
A.~Guzhov, F.~Raue, J.~Hees, and A.~Dengel, ``Audioclip: Extending clip to image, text and audio,'' in \emph{ICASSP 2022-2022 IEEE International Conference on Acoustics, Speech and Signal Processing (ICASSP)}.\hskip 1em plus 0.5em minus 0.4em\relax IEEE, 2022, pp. 976--980.

\bibitem{huang2022mulan}
Q.~Huang, A.~Jansen, J.~Lee, R.~Ganti, J.~Y. Li, and D.~P.~W. Ellis, ``Mulan: {A} joint embedding of music audio and natural language,'' in \emph{Proceedings of the 23rd International Society for Music Information Retrieval Conference}, P.~Rao, H.~A. Murthy, A.~Srinivasamurthy, R.~M. Bittner, R.~C. Repetto, M.~Goto, X.~Serra, and M.~Miron, Eds., 2022, pp. 559--566.

\bibitem{donahue2023singsong}
C.~Donahue, A.~Caillon, A.~Roberts, E.~Manilow, P.~Esling, A.~Agostinelli, M.~Verzetti, I.~Simon, O.~Pietquin, N.~Zeghidour \emph{et~al.}, ``Singsong: Generating musical accompaniments from singing,'' \emph{arXiv preprint arXiv:2301.12662}, 2023.

\bibitem{zhang2023llama}
R.~Zhang, J.~Han, A.~Zhou, X.~Hu, S.~Yan, P.~Lu, H.~Li, P.~Gao, and Y.~Qiao, ``Llama-adapter: Efficient fine-tuning of language models with zero-init attention,'' \emph{arXiv preprint arXiv:2303.16199}, 2023.

\bibitem{van2017neural}
A.~Van Den~Oord, O.~Vinyals \emph{et~al.}, ``Neural discrete representation learning,'' \emph{Advances in neural information processing systems}, vol.~30, 2017.

\bibitem{defossez2022high}
A.~D{\'e}fossez, J.~Copet, G.~Synnaeve, and Y.~Adi, ``High fidelity neural audio compression,'' \emph{arXiv preprint arXiv:2210.13438}, 2022.

\bibitem{wu2024music}
S.-L. Wu, C.~Donahue, S.~Watanabe, and N.~J. Bryan, ``Music controlnet: Multiple time-varying controls for music generation,'' \emph{IEEE/ACM Transactions on Audio, Speech, and Language Processing}, vol.~32, pp. 2692--2703, 2024.

\bibitem{devlin2018bert}
J.~D. M.-W.~C. Kenton and L.~K. Toutanova, ``Bert: Pre-training of deep bidirectional transformers for language understanding,'' in \emph{Proceedings of NAACL-HLT}, 2019, pp. 4171--4186.

\bibitem{radford2019language}
A.~Radford, J.~Wu, R.~Child, D.~Luan, D.~Amodei, I.~Sutskever \emph{et~al.}, ``Language models are unsupervised multitask learners,'' \emph{OpenAI blog}, vol.~1, no.~8, p.~9, 2019.

\bibitem{li2021prefix}
X.~L. Li and P.~Liang, ``Prefix-tuning: Optimizing continuous prompts for generation,'' in \emph{Proceedings of the 59th Annual Meeting of the Association for Computational Linguistics and the 11th International Joint Conference on Natural Language Processing (Volume 1: Long Papers)}, 2021, pp. 4582--4597.

\bibitem{liu2021p}
X.~Liu, K.~Ji, Y.~Fu, W.~Tam, Z.~Du, Z.~Yang, and J.~Tang, ``P-tuning: Prompt tuning can be comparable to fine-tuning across scales and tasks,'' in \emph{Proceedings of the 60th Annual Meeting of the Association for Computational Linguistics (Volume 2: Short Papers)}, 2022, pp. 61--68.

\bibitem{hu2021lora}
E.~J. Hu, P.~Wallis, Z.~Allen-Zhu, Y.~Li, S.~Wang, L.~Wang, W.~Chen \emph{et~al.}, ``Lora: Low-rank adaptation of large language models,'' in \emph{International Conference on Learning Representations}, 2021.

\bibitem{gao2023llama}
P.~Gao, J.~Han, R.~Zhang, Z.~Lin, S.~Geng, A.~Zhou, W.~Zhang, P.~Lu, C.~He, X.~Yue \emph{et~al.}, ``Llama-adapter v2: Parameter-efficient visual instruction model,'' \emph{arXiv preprint arXiv:2304.15010}, 2023.

\bibitem{zeghidour2021soundstream}
N.~Zeghidour, A.~Luebs, A.~Omran, J.~Skoglund, and M.~Tagliasacchi, ``Soundstream: An end-to-end neural audio codec,'' \emph{IEEE/ACM Transactions on Audio, Speech, and Language Processing}, vol.~30, pp. 495--507, 2021.

\bibitem{jiang2019large}
J.~Jiang, K.~Chen, W.~Li, and G.~Xia, ``Large-vocabulary chord transcription via chord structure decomposition,'' in \emph{Proceedings of the 20th International Society for Music Information Retrieval Conference}, A.~Flexer, G.~Peeters, J.~Urbano, and A.~Volk, Eds., 2019, pp. 644--651.

\bibitem{wu2021omnizart}
Y.-T. Wu, Y.-J. Luo, T.-P. Chen, I.~Wei, J.-Y. Hsu, Y.-C. Chuang, L.~Su \emph{et~al.}, ``Omnizart: A general toolbox for automatic music transcription,'' \emph{The Journal of Open Source Software}, vol.~6, no.~68, p. 3391, 2021.

\bibitem{bock2016madmom}
S.~B{\"o}ck, F.~Korzeniowski, J.~Schl{\"u}ter, F.~Krebs, and G.~Widmer, ``Madmom: A new python audio and music signal processing library,'' in \emph{Proceedings of the 24th ACM international conference on Multimedia}, 2016, pp. 1174--1178.

\bibitem{rafii2017musdb18}
Z.~Rafii, A.~Liutkus, F.-R. St{\"o}ter, S.~I. Mimilakis, and R.~Bittner, ``Musdb18-a corpus for music separation,'' 2017.

\bibitem{Wu2021}
\BIBentryALTinterwordspacing
Y.-T. Wu, Y.-J. Luo, T.-P. Chen, I.-C. Wei, J.-Y. Hsu, Y.-C. Chuang, and L.~Su, ``Omnizart: A general toolbox for automatic music transcription,'' \emph{Journal of Open Source Software}, vol.~6, no.~68, p. 3391, 2021. [Online]. Available: \url{https://doi.org/10.21105/joss.03391}
\BIBentrySTDinterwordspacing

\bibitem{goto2002rwc}
M.~Goto, H.~Hashiguchi, T.~Nishimura, and R.~Oka, ``Rwc music database: Popular, classical and jazz music databases,'' in \emph{Proceedings of the 3rd International Society for Music Information Retrieval Conference}, 2002.

\bibitem{defossez2019demucs}
A.~D{\'e}fossez, N.~Usunier, L.~Bottou, and F.~Bach, ``Demucs: Deep extractor for music sources with extra unlabeled data remixed,'' \emph{arXiv preprint arXiv:1909.01174}, 2019.

\bibitem{wu2023large}
Y.~Wu, K.~Chen, T.~Zhang, Y.~Hui, T.~Berg-Kirkpatrick, and S.~Dubnov, ``Large-scale contrastive language-audio pretraining with feature fusion and keyword-to-caption augmentation,'' in \emph{ICASSP 2023-2023 IEEE International Conference on Acoustics, Speech and Signal Processing (ICASSP)}.\hskip 1em plus 0.5em minus 0.4em\relax IEEE, 2023, pp. 1--5.

\bibitem{kilgour2018fr}
K.~Kilgour, M.~Zuluaga, D.~Roblek, and M.~Sharifi, ``Fr\'echet audio distance: A metric for evaluating music enhancement algorithms,'' \emph{arXiv preprint arXiv:1812.08466}, 2018.

\end{thebibliography}

%
%
%
%
%

\end{document}